\documentclass[conference]{IEEEtran}
\usepackage{times}

\usepackage[numbers]{natbib}
\usepackage{multicol}
\usepackage[bookmarks=true, hidelinks]{hyperref}

\usepackage{mathtools}
\usepackage{amsmath,amssymb}
\usepackage{graphicx}
\usepackage{svg}
\usepackage{booktabs}
\usepackage[utf8]{inputenc}
\usepackage{pgfplots}
\usepackage{tabularx}
\usepackage{xcolor, colortbl}
\usepackage[T1]{fontenc}
\usepackage{multirow}
\DeclareUnicodeCharacter{2212}{−}
\usepgfplotslibrary{groupplots,dateplot}
\usetikzlibrary{patterns,shapes.arrows}
\pgfplotsset{compat=newest}

\definecolor{lightgray}{gray}{0.9}

\DeclareMathOperator{\tr}{tr}

\newcommand{\specialcell}[2][c]{%
\begin{tabular}[#1]{@{}c@{}}#2\end{tabular}}
\newcommand\shrink{\hspace{0em}}

\makeatletter
\newcommand*{\transpose}{%
  {\mathpalette\@transpose{}}%
}
\newcommand*{\@transpose}[2]{%
  \raisebox{\depth}{$\m@th#1\intercal$}%
}
\makeatother

\usepackage{etoolbox}
\makeatletter
\patchcmd{\@makecaption}
  {\scshape}
  {}
  {}
  {}
\patchcmd{\@makecaption}
  {\\}
  {.\ }
  {}
  {}
 \patchcmd{\@makecaption}
  {\raggedright}
  {}
  {}
  {}
\makeatother
\begin{document}

\title{End-to-End Learning of Hybrid Inverse Dynamics Models for Precise and Compliant Impedance Control}




%
\author{\authorblockN{Moritz Reuss\authorrefmark{1}\authorrefmark{3},
Niels van Duijkeren\authorrefmark{1},
Robert Krug\authorrefmark{1},
Philipp Becker\authorrefmark{4},
Vaisakh Shaj\authorrefmark{4}\authorrefmark{2} and
Gerhard Neumann\authorrefmark{4}}
\authorblockA{\authorrefmark{1}Bosch Corporate Research, Renningen, Germany}
\authorblockA{\authorrefmark{2}LCAS, 
University Of Lincoln, UK}
\authorblockA{\authorrefmark{3}Intuitive Robots Lab, 
Karlsruhe Institute of Technology, Germany}
\authorblockA{\authorrefmark{4}Autonomous Learning Robots, 
Karlsruhe Institute of Technology, Germany}}

\maketitle

\begin{abstract}

It is well-known that inverse dynamics models can improve tracking performance in robot control. 
These models need to precisely capture the robot dynamics, which consist of well-understood components, e.g., rigid body dynamics, and effects that remain challenging to capture, e.g., stick-slip friction and mechanical flexibilities. Such effects exhibit hysteresis and partial observability, rendering them, particularly challenging to model. 
Hence, hybrid models, which combine a physical prior with data-driven approaches are especially well-suited in this setting.
We present a novel hybrid model formulation that enables us to identify fully physically consistent inertial parameters of a rigid body dynamics model which is paired with a recurrent neural network architecture, allowing us to capture unmodeled partially observable effects using the network memory. 
We compare our approach against state-of-the-art inverse dynamics models on a 7 degree of freedom manipulator. 
Using data sets obtained through an optimal experiment design approach, we study the accuracy of offline torque prediction and generalization capabilities of joint learning methods.
In control experiments on the real system, we evaluate the model as a feed-forward term for impedance control and show the feedback gains can be drastically reduced to achieve a given tracking accuracy. 
\end{abstract}

\IEEEpeerreviewmaketitle

\section{Introduction}
Automating industrial tasks such as shaft insertion or cable plugging in semi-structured environments remains challenging for industrial robots.
Simultaneous compliant and precise motion tracking is required to prevent the damage of fragile parts in assembly steps while maintaining precise movements. 
However, achieving compliant behavior and precise motion are typically conflicting objectives in feedback control.
Feedback controllers can provide precise movements with high feedback gains but this renders the manipulator stiff and therefore non-compliant.
By using a precise inverse model of the robot dynamics
the reliance on the feedback controller for tracking accuracy can be reduced.
Using such a feed-forward compensator enables simultaneous compliant and precise tracking.
Identification of a precise inverse dynamics model of an industrial robot is a key research area in precise and compliant control. 
A classical approach to derive the equations of motion of the rigid body dynamics is the Recursive Newton-Euler Algorithm.
These models have the advantage of being valid in the whole state space while having low computational complexity and are used widely in robotics.
However, they are unable to capture non-rigid and temporal effects. 
Black-box models such as neural networks offer efficient data-driven optimization with a low bias.
However, such purely data-driven models are typically data inefficient, lack any stability guarantees for extrapolation, and
only achieve good performance in the vicinity of the training data.
These drawbacks disallow their usage for safety-critical applications in semi-structured environments \cite{siciliano2009modelling}. 

In the last few years structured models, grey-box models, and hybrid models have gained popularity in the research community \cite{geist2021structured}.
They combine the physical prior of rigid-body models with efficient data-driven optimization to complement their individual strengths. 
Combining the generalization of rigid-body dynamics models with additional data-driven terms to further optimize task-specific accuracy makes residual hybrid models ideal for compliant and precise motion tracking. 

Previous residual hybrid model implementations rely on stateless feed-forward neural networks or Gaussian Processes to learn the error of the rigid body dynamics model. 
Stateless models are limited in their ability to capture partially observable dynamic effects, such as hysteresis and body elasticities. 
Utilizing recurrent neural networks such as Long Short-Term Memory networks (LSTM) \cite{hochreiter1997long} has the potential to capture these effects and improve the model accuracy.

Hybrid robot models that combine rigid-body dynamics, parameterized in the inertial properties of the kinematic chain, with a residual DNN are most easily obtained in a sequential identification process.
Using the established toolchains for each step, the inertial parameters are matched with the collected data first, then the residual DNN is trained on the mismatch.
Yet, combining the learning of the inertia parameters and the residual DNN in one step can potentially improve the obtained hybrid model.
As the residual DNN captures the non-rigid-body dynamics effects increasingly well, the inertia parameters will fit on a less biased residual error \cite{geist2021structured}.

Recently, a differentiable version of the Recursive-Newton-Euler algorithm (DiffNEA) was introduced in \cite{sutanto2020encoding, lutterDifferentiableNewtonEulerAlgorithm2021}.
DiffNEA is optimized using gradient descent methods to find valid inertial parameters, which can be used for end-to-end learning of inverse dynamics.
First results in end-to-end learning of forward dynamics show the potential of applying it to inverse dynamics learning \cite{lutterDifferentiableNewtonEulerAlgorithm2021}.

In this work, we extend previous ideas of end-to-end learning to inverse dynamics learning of hybrid models and target the challenges of the system identification process.
We propose a new residual hybrid model architecture consisting of a rigid-body model with a simplified Coulomb friction model and an LSTM as its residual model.
We evaluate the performance of the joint learned residual hybrid models to accurately predict the inverse dynamics of a Franka Emika Panda Robot from offline data.
Further, we introduce an alternative formulation of DiffNEA as an unconstrained nonlinear regression problem to identify fully physically consistent link frame inertial parameters, which we name "DiffBary".
Both methods will be used for end-to-end learning of hybrid inverse dynamics models.
We compare our model against recent state-of-the-art structured and unstructured model architectures in terms of their offline torque prediction accuracy and generalization capabilities and show superior performance by our hybrid architecture.
Motion tracking experiments are conducted with a real robot arm to evaluate different models inside a joint state impedance controller.
The effectiveness of the joint hybrid model training is validated through observed performance improvements of an impedance controller for a 7-DoF robot arm setup and offline evaluation of experiment data.
Our results show that using our residual hybrid model enables simultaneous compliant and precise motion tracking.

\section{Background and Related Work}

The goal of inverse dynamics modeling is to find the function $f_{\theta}$,
\begin{equation}
    \boldsymbol{\tau}^{t}_{\text{ref}} = f_{\theta}(\boldsymbol{q}^{t}, \boldsymbol{\dot{q}}^{t}, \boldsymbol{\ddot{q}}_{\text{des}}^{t})
\end{equation}
that best maps the current robot joint angles $\boldsymbol{q}^{t}$, velocities $\boldsymbol{\dot{q}}^{t}$ and desired accelerations $\boldsymbol{\ddot{q}}_{\text{des}}^{t}$ to the torque necessary to achieve the desired acceleration.
The model is optimized to minimize the mean squared error (MSE) between the torques predicted by the model and the ground truth from a data set consisting of numerous trajectories  $\mathcal{D} = \{ \{(\boldsymbol{q}^{t}, \boldsymbol{\dot{q}}^{t}, \boldsymbol{\ddot{q}}_{\text{des}}^{t}), \boldsymbol{\tau}^{t}_{ref}\}_{t=1}^{T} \}_{n=1}^{N}$, where $T$ refers to the number of time steps and $N$ to the number of trajectories.
For a single trajectory the loss is defined as:
\begin{equation}\label{eq:cost_function}
    \mathcal{L}_{MSE} = \frac{1}{T} \sum_{t=1}^{T} \left\| \boldsymbol{\tau}^{t}_{\text{ref}}- f_{\theta}(\boldsymbol{q}^{t}, \boldsymbol{\dot{q}}^{t}, \boldsymbol{\ddot{q}}_{\text{des}}^{t}, \boldsymbol{\theta}) \right\|^{2}
\end{equation}
In this work, we consider three different model types for $f_{\theta}$: classical rigid-body dynamics models (RBD), black-box models, and hybrid models that combine rigid-body dynamics with black-box models.

\subsection{Rigid-body Dynamics Models}
The robot manipulator is described as a chain of rigid bodies, their equations of motion (EOMs) in generalized coordinates can be written as
\begin{equation}\label{eom}
    \boldsymbol{M}(\boldsymbol{q}^{t}) \boldsymbol{\ddot{q}}^{t} + \boldsymbol{V}(\boldsymbol{q}^{t},\boldsymbol{\dot{q}}^{t}) + \boldsymbol{G}(\boldsymbol{q}^{t}) = \boldsymbol{\tau}^{t}_{\text{ext}}+\boldsymbol{\tau}^{t}_{\text{ref}}\end{equation}
where $\boldsymbol{M}(\boldsymbol{q})$ denotes the mass matrix, $ \boldsymbol{V}(\boldsymbol{q},\boldsymbol{\dot{q}})$ models the Coriolis and centrifugal effects, $\boldsymbol{G}(\boldsymbol{q})$ models the effect of gravity, $\boldsymbol{\tau}^{t}_{\text{ext}}$ denotes the external (exogenous) torques, and $\boldsymbol{\tau}^{t}_{\text{ref}}$ the actuation torques.  
For free space motion, in absence of contact, it can be considered that $\boldsymbol{\tau}^{t}_{\text{ext}} = \boldsymbol{0}$.
For inverse dynamics identification, the linearity of the EOMs in the so-called barycentric parameters is commonly exploited \cite{sousa2014physical}:
\begin{equation}
\label{eq:linear EOM}
    \boldsymbol{\tau}^{t}_{\text{ref}} = f_{\text{RBD}}(\boldsymbol{q}^{t}, \boldsymbol{\dot{q}}^{t}, \boldsymbol{\ddot{q}}^{t}) = \boldsymbol{Y}(\boldsymbol{q}^{t}, \boldsymbol{\dot{q}}^{t}, \boldsymbol{\ddot{q}}^{t})\boldsymbol{\theta}
\end{equation}
where the matrix $\boldsymbol{Y}(\boldsymbol{q}^{t}, \boldsymbol{\dot{q}}^{t}, \boldsymbol{\ddot{q}}^{t})$ only depends on the robot kinematic properties and $\boldsymbol{\theta} \in \mathbf{R}^{10k}$ is a parameter vector that contains the values $\boldsymbol{\theta}_{k}=[m_{k}, l_{k,x}, l_{k,y}, l_{k,z}, L_{k,xx}, L_{k,xy}, L_{k,xz}, L_{k,yy}, L_{k,yz}, L_{k,zz}]^{\transpose}$ for each link $k$.
The set of parameters consists of the mass $m_{k}$, the first mass moment $\boldsymbol{l}_{k}$ and the six elements characterizing the symmetric inertia tensor $\boldsymbol{L}_{k}$; where the latter is the inertia expressed in the link frame, which is in the joint connecting link $k$ to link $k-1$ \cite{atkeson1986estimation}.
The inertia tensor in the center of mass $\boldsymbol{I}_{k}$ and the link frame inertia tensor $\boldsymbol{L}_{k}$ are related by the parallel axis theorem 
\begin{equation}
\label{eq: L to I}
    \boldsymbol{L}_{k} = \boldsymbol{I}_{k} - m_{k} \boldsymbol{S}(\boldsymbol{r}_{k}) \boldsymbol{S}(\boldsymbol{r}_{k}) = \boldsymbol{I}_{k} + \frac{1}{m_{k}}\boldsymbol{S}(\boldsymbol{l}_{k})^{\transpose}\boldsymbol{S}(\boldsymbol{l}_{k}),
\end{equation}
where  $\boldsymbol{l}_{k}=m_{k}\boldsymbol{r}_{k}$, with $\boldsymbol{r}_{k}$ the position of the center of mass of link $k$ in the link frame, and $\boldsymbol{S} \colon \mathbf{R}^3 \to \mathbf{R}^{3 \times 3}$ the skew symmetric matrix operator such that $\boldsymbol{a} \times \boldsymbol{b} = \boldsymbol{S}(\boldsymbol{a}) \boldsymbol{b}$ for all $\boldsymbol{a}, \boldsymbol{b} \in \mathbf{R}^3$.
By linearity of Eq. \eqref{eq:linear EOM}, one can estimate the inertial parameters using ordinary least square methods \cite{atkeson1986estimation}.
Despite nonlinearity of the matrix $\boldsymbol{Y}(\boldsymbol{q}^{t}, \boldsymbol{\dot{q}}^{t}, \boldsymbol{\ddot{q}}^{t})$ from Eq. \eqref{eq:linear EOM} in its arguments, its columns are not generally linearly independent.
In order to ensure Eq. \eqref{eq:cost_function} has a unique minimizer, the estimation usually aims at finding a set of "base parameters" \cite{mayeda1988base} instead.

Additional constraints restrict the values of the inertial parameters. 
\citet{yoshida2000verification} first introduced the conditions for physically consistent inertial parameters, 
\begin{equation}
\label{pc_condition}
    \begin{cases}
      m_k>0\\
      \boldsymbol{I}_k \succ \boldsymbol{0}
    \end{cases} 
\end{equation}
which states, that all masses need to be positive and the inertia tensors positive definite.

\citet{traversaro2016identification} highlighted, that those two conditions are not sufficient to guarantee fully physically consistent inertial parameters. 
The eigenvalues of the inertia tensor must fulfill the so-called triangle inequalities.
The inertia tensor can be expressed by the principal axes of rotation and the according principal moments of inertia $Y_{X},Y_{Y},Y_{Z}$.
One can decompose the inertia tensor,
\begin{equation}
\label{eq: inertia tensor relation}
    \boldsymbol{I}_{k} = \boldsymbol{R} \boldsymbol{Y}_{k} \boldsymbol{R}^{\transpose}
\end{equation}
in a rotation matrix $\boldsymbol{R}$ and a diagonal matrix $\boldsymbol{Y}_{k}$ with the principal moments of inertia on the diagonal.
These must be positive and must fulfil the following inequalities
\begin{equation}
\label{tri-ineq}
    \begin{cases}
      Y_{X} + Y_{Y} > Y_{Z}\\
      Y_{Y} + Y_{Z} > Y_{X}\\
      Y_{X} + Y_{Z} > Y_{Y}
    \end{cases} 
\end{equation}
to enforce a positive mass density.
A set of inertial parameters is only fully physically consistent, if all three conditions in Eq.  \eqref{pc_condition} and Eq. \eqref{tri-ineq} are satisfied.
In \cite{traversaro2016identification} a manifold optimization method is presented to estimate fully physically consistent parameters.

\citet{wensing2017linear} have formulated the linear inequalities (LMI) in dependency of the rotational inertia matrix $\boldsymbol{I}_{C}$, which is required to have a positive-definite covariance of its mass distribution.
\begin{equation}
\label{eq:LMI COG}
    \Sigma_{C} = \frac{1}{2} \tr(\boldsymbol{I}_{C}) \boldsymbol{1}_{3} - \boldsymbol{I}_{C} \succ \boldsymbol{0}
\end{equation}
$\boldsymbol{1}_3$ denotes a 3x3 identity matrix and $\tr$ the trace operator, which takes the sum of the diagonal elements of a matrix.
The above linear matrix inequality is shown to be equivalent to the inequalities in Eq. \eqref{tri-ineq}.

\citet{wensing2017linear} and \citet{sousa2019inertia} independently showed that all three conditions in Eq. \eqref{eq: inertia tensor relation} and Eq. \eqref{tri-ineq} can be combined into a single LMI per robot link.
\begin{equation}
\label{eq:LMI inertial paraeters}
    \boldsymbol{S}_k = \begin{bmatrix}
    \frac{\tr(\boldsymbol{L}_{k})}{2} \boldsymbol{1}_{3} - \boldsymbol{L}_{k} & \boldsymbol{l}_{k} \\
    \boldsymbol{l}_{k}^{\transpose}  & m_{k}
    \end{bmatrix}
    \succ \boldsymbol{0}.
\end{equation}
The LMIs can be incorporated in a semi-definite programming (SDP) problem that minimizes Eq. \eqref{eq:cost_function}.

\subsection{Hybrid Inverse Dynamics Models}

Several types of hybrid or grey-box models have been introduced in recent years.
Physics-inspired neural networks such as Deep Lagriangen Networks~(DeLaN) \cite{lutter2019deep} or Hamiltonian networks \cite{greydanus2019hamiltonian} guarantee physically plausible dynamic models, which conserve energy.
They incorporate structure of the system dynamics into the neural network architecture, are limited to model conservative forces, and use additional neural networks to capture all non-conservative effects.
Residual hybrid models combine a rigid-body model with an additional error model to learn the error of the RBD model part \cite{gupta2020structured}.
For the residual model one can use either Gaussian processes~(GPs) or neural networks~(NNs).
\citet{nguyen2010using} and \cite{meier2016towards} used GP models to learn the error of an inverse dynamics model.
Several recent works combine a standard Multilayer perceptron~(MLP) neural network with a rigid-body model \cite{hitzler2019learning, kappler2017new, lutterDifferentiableNewtonEulerAlgorithm2021}.
A similar approach has been applied to physics-inspired neural networks to capture the non-conservative forces \cite{gupta2020structured}.
Previous work successfully applied different recurrent neural networks for black-box inverse dynamics models \cite{rueckert2017learning, shaj2020action}.
However, to the best of our knowledge, there is no previous work in applying recurrent models as residual models in inverse dynamics.

The Differentiable Newton Euler Algorithm~(DiffNEA) implements a recursive Newton-Euler algorithm that uses virtual parameters to guarantee physical consistency of the inertial parameters during training \cite{sutanto2020encoding, lutterDifferentiableNewtonEulerAlgorithm2021}.
A similar implementation has been presented in \cite{ledezma2017first}.
The structure of the fully physically conditions from Eq. \eqref{pc_condition} and Eq. \eqref{eq:LMI COG} are incorporated by using a set of unbounded virtual parameters to learn physically consistent inertial parameters with gradient descent. 
\citet{lutterDifferentiableNewtonEulerAlgorithm2021} applied the DiffNEA algorithm with additional actuator models in forward dynamics learning to capture dissipating forces.
They trained the parameters jointly and achieved similar results to models learned in separate optimizations for systems up to 4 degrees of freedom. 
Our work is inspired by the method of \citet{sutanto2020encoding} to incorporate the physical conditions into the optimization formulation. 
We introduce an alternative formulation, which uses a set of unbounded virtual parameters to directly learn a set of fully physically consistent link frame inertial parameters for Eq. \eqref{eq:linear EOM}.
We show that both formulations are equally suited for joint residual hybrid model learning.

\section{End-to-end Learning of Hybrid Inverse Dynamics Models}
\label{sec: Methodology}
\begin{figure}
    \centering
    \includegraphics[width=0.49\textwidth]{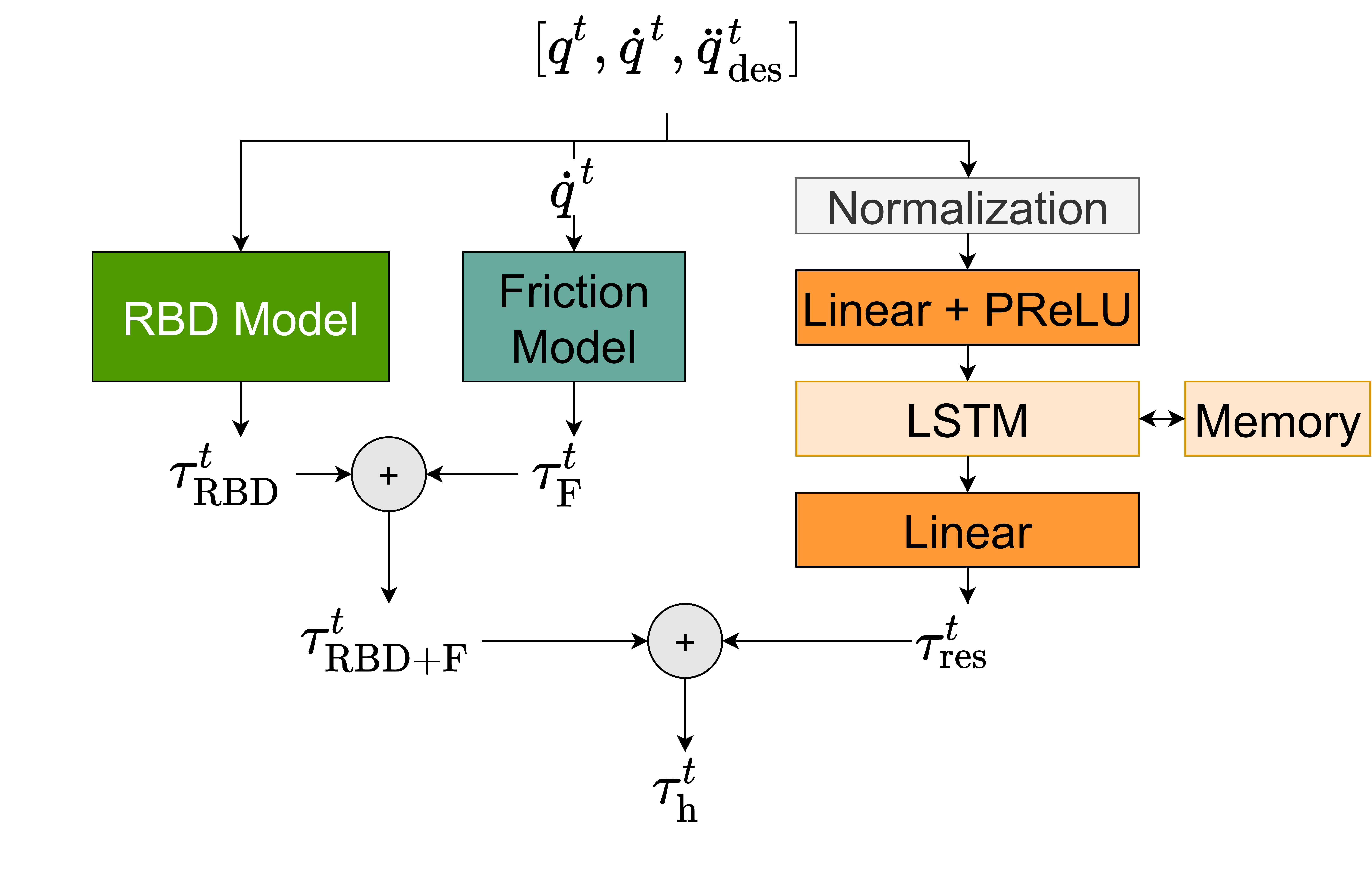}
    \caption{Schematic of the residual hybrid model architecture proposed in section \ref{sec: Methodology} consisting of the rigid-body, the sign friction function (Eq. \eqref{eq: sign friction}) and the LSTM error model, which has an additional input normalization layer. The final output torque sum $\tau_H^t$ is normalized for training.}
    \label{fig: structured model overview}
\end{figure}
Our hybrid architecture pairs a rigid body dynamics model with a recurrent neural network (RNN) which models the residuals of the RBD model and accounts for uncaptured effects such as the flexibility of the robot. 
Our proposed architecture consists of three components, 
\begin{equation}
    \boldsymbol{\tau}^{t}_{\text{ref}} = f_{\text{RBD}}(\boldsymbol{q}^{t}, \boldsymbol{\dot{q}}^{t}, \boldsymbol{\ddot{q}}_{\text{des}}^{t}) + f_{\text{F}}(\boldsymbol{\dot{q}}^{t}) + f_{\text{RNN}}(\boldsymbol{q}^{t}, \boldsymbol{\dot{q}}^{t}, \boldsymbol{\ddot{q}}_{\text{des}}^{t}, \boldsymbol{\theta}_{\text{RNN}}),
\end{equation}
where $f_{\text{RBD}}$ is a traditional rigid-body model using the Equations of Motion (EOMs) from Eq. \eqref{eq:linear EOM}, $f_{\text{F}}$ is a friction model, and $f_{\text{RNN}}$ an RNN which can capture partially observable effects such as joint and link flexibilities and more complex friction effects.
A general overview of the structured model is visualized in Figure \ref{fig: structured model overview}.

\subsection{Recurrent Residual Models}
\label{sec: Residual Neural Network}
While the learned rigid-body dynamics model is globally valid, it is only able to capture rigid and stateless effects.
By extending the RBD model with an additional error model, we can increase the accuracy of the model and also capture unmodeled effects such as slip-stick friction. While previous approaches rely on standard MLPs to capture the residual of the RBD model, we argue that recurrent models are better suited to capture these unmodeled effects as most of these effects have an internal, unobserved state that needs to be inferred \cite{johanastrom2008revisiting}. 
For our implementation we use Long Short-Term Memory networks (LSTM) \cite{hochreiter1997long}.
LSTMs are commonly used in dynamic system modeling and have been successfully applied to black-box inverse dynamics model learning of a 7-DOF robot arm in \cite{rueckert2017learning} using offline data.
Yet, as we are also interested in deploying our architecture on the real robot in real-time (i.e., $1\textrm{kHz}$), we have to comply with tight computation time constraints and therefore resort to rather small network architectures. Our proposed residual network architecture consists of three layers: a single linear layer with PReLU activation functions, an LSTM layer, and an output layer consisting of seven neurons without any activation function to predict the seven individual residual joint torque values.
We introduce additional normalization layers to normalize the input data for the neural network with predefined values.
In between the model layers, we apply the layer normalization from \cite{ba2016layer}.
Layer normalization is known to reduce the covariance shift between train and test set and boosts training speed and generalization.
For comparison, we also used a 3-layer MLP network to show the effectiveness of the LSTM at capturing temporal effects in the inverse dynamics data.
Together with the fully differentiable DiffBary and DiffNEA methods (see next sub-section), we learn the parameters of the residual LSTM alongside the rigid body dynamics parameters and the parameters of the friction model.

\subsection{Learning Physically Consistent Barycentric Parameters}
\label{sec: Learning fully physically consistent}
To allow gradient-based end-to-end learning, we require an unconstrained parameterization of the rigid body dynamics model that always yields physically consistent parameters. 
To this end, we introduce an alternative formulation to DiffNEA \cite{sutanto2020encoding} called "Differentiable Barycentric" (DiffBary).
The method also uses a set of virtual unbounded parameters and computes the inertial parameters from the virtual parameters. 
Further, we apply this idea directly to the linear equation of motion with the barycentric parameters.
We construct the LMI matrix $\boldsymbol{S}_{k}$ of Eq. \eqref{eq:LMI inertial paraeters}, which expresses the conditions for the physical consistency from the virtual unbounded parameters and afterward retrieve the inertial parameters from it.
By ensuring the positive definiteness of the matrix $\boldsymbol{S}_{k}$ of Eq. \eqref{eq:LMI inertial paraeters}, we can guarantee that all physical conditions of the inertial parameters for the link $k$, Eq. \eqref{pc_condition} and Eq. \eqref{tri-ineq}, are satisfied.
Inspired by the idea of DiffNEA, we learn the terms of the Cholesky decomposition of $\boldsymbol{S}_{k}$ and thereby enforce positive-definiteness of the matrix itself, i.e., 
for every generic link frame $k$, we introduce a lower triangular matrix $\boldsymbol{A}_{k}$ with 10 unbounded virtual parameters.
The product 
$\boldsymbol{A}_{k}\boldsymbol{A}_{k}^{\transpose}$ together with a small positive bias $r=10^{-8}$ on the diagonal entries guarantees strict positive-definiteness: 
\begin{equation}
    \boldsymbol{S}_{k} = \boldsymbol{A}_{k}\boldsymbol{A}_{k}^{\transpose} + r\boldsymbol{1}_{4}.
\end{equation}
Now, all inertial parameters are reconstructed from this matrix $\boldsymbol{S}_{k}$. 
We refer to Eq. \eqref{eq: S from unbounded} and Eq. \eqref{eq: retrive bary params} in the appendix for detailed information on how to retrieve the individual parameters.
Next, we can directly use Eq. \eqref{eq:linear EOM} to calculate the torques depending on the current system state.

As we can now guarantee fully physically consistent inertial parameters without additional constraints,
this method can be straightforwardly applied for end-to-end learning of residual hybrid models or we can use it separately to identify inertial parameters.
Our parameterization of the rigid body dynamics model allows us to learn physically consistent inertial parameters alongside the neural network weights using stochastic gradient descent algorithms such as Adam \cite{kingma2015adam}.

\subsection{Friction Model}
For the friction model, we employ a modified Coulomb model \cite{olsson1998friction}, which is only dependent on the joint velocity,
\begin{equation}
\label{eq: sign friction}
    f_{\text{F,k}}(\dot{q}^{t}_{k}) =
\left\{
        \begin{array}{ll}
                 \sigma_{C,k} \text{sign}(\dot{q}^{t}_{k})  & \mbox{if } |\dot{q}^{t}_{k} | > \dot{q}^{t}_{\text{lim},k} \\
                \sigma_{C,k} \dot{q}^{t}_{k} / \dot{q}^{t}_{\text{lim},k} & \mbox{if } |\dot{q}^{t}_{k} | \leq \dot{q}^{t}_{\text{lim},k}
        \end{array}
\right.
\end{equation}
where $\sigma_{C,k}$ is a friction coefficient and $\dot{q}^{t}_{\text{lim},k} = 0.02$ for every degree of freedom $k$.
The friction terms appear linearly in the joint torques, we therefore extend our initial formulation of the EOMs in Eq. \eqref{eq:linear EOM}, to include the friction functions and add the parameters to our link parameter vector $\boldsymbol{\theta}_{k}=[m_{k}, \boldsymbol{l}_{k},\boldsymbol{L}_{k}, \sigma_{C,k}]^{\transpose} \in \mathbf{R}^{k \times 11} $

\section{Offline Evaluation}
\label{sec: Evaluation}
We evaluate our end-to-end learned structured inverse dynamics model shown in Figure \ref{fig: structured model overview} on a Franka Emika Panda robot and compare it to several rigid-body dynamics, black-box, and hybrid baseline models. 
First, we evaluate the approaches on their offline prediction performance. 
We compare a stand-alone version of DiffBary to other approaches for rigid body dynamics identification.
Further, we analyze how the joint training of our residual hybrid architecture disentangles the RBD, friction, and residual components and investigate the generalization capabilities of the different models by explicit evaluation on out-of-distribution test data.
Subsequently, we deploy the approaches on the real robot and evaluate their performance in motion tracking experiments using real-time impedance control on the real robot. 
We test the joint angle tracking accuracy of the different approaches using different controller feedback gains and show that structured models are suited for compliant and precise motion control.

\subsection{Baselines}
\label{sec:baseline}
\begin{figure*}[t]
    \centering
    \includegraphics[width=1\textwidth]{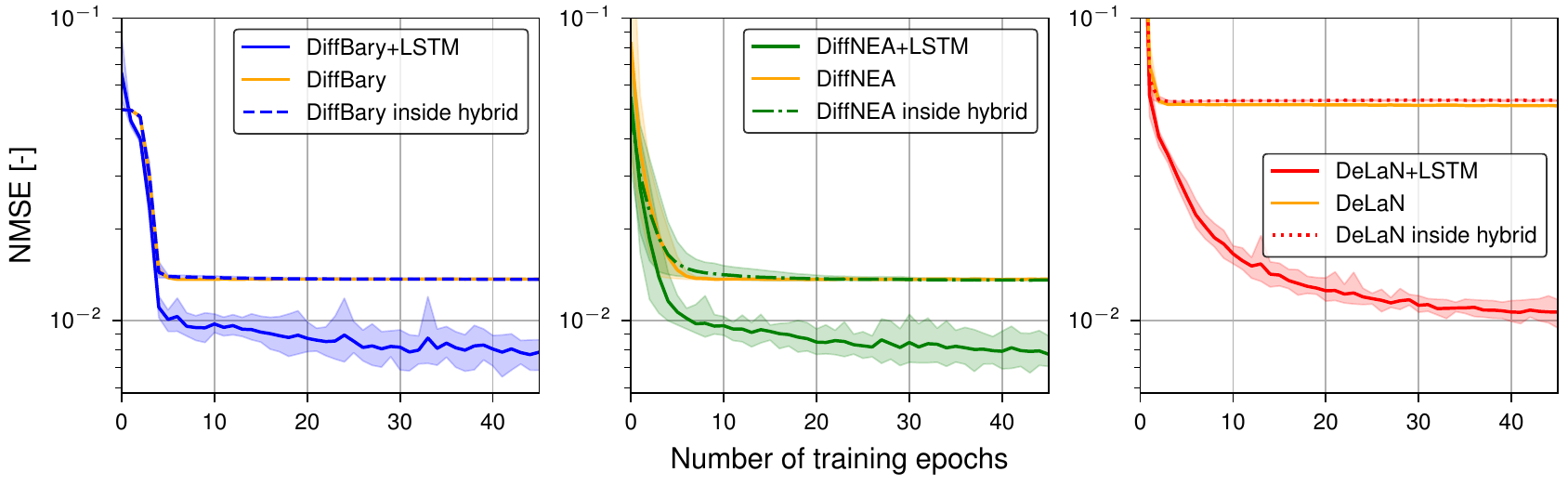}
    \caption{Comparison of end-to-end inverse dynamics learning on a small system identification data set. All 3 rigid-body models are trained individually and combined with a residual LSTM as a hybrid model. The validation loss of the rigid-body models inside the hybrid model is monitored and visualized to compare the end-to-end learning efficiency of each model type. DiffNEA and DiffBary inside the hybrid model converge to the same solution after a few epochs for all 10 different seeds. The DeLaN model can also be learned jointly with an LSTM. All 3 models achieve a similar loss value if learned independently or in an end-to-end setting with the residual LSTM.}
    \label{fig: joint learning comparison}
\end{figure*}
We compare our proposed hybrid model architecture with the following baselines. 

\textbf{Rigid Body Dynamics Baselines}: 
We use a classical rigid-body dynamics baseline with fully physically consistent inertial parameters obtained using SDP \cite{sousa2019inertia} from system identification literature.

\textbf{Hybrid Model Baselines}: We use the DiffNEA algorithm \cite{sutanto2020encoding} and the DeLaN model \cite{lutter2019deep} as our hybrid physics-inspired neural network baselines. Further, we use a residual hybrid model with an MLP as a baseline for the residual LSTM.

\textbf{Black-box Baselines}: 
We also compare it with a black-box LSTM model.
The black-box LSTM has the same architecture as our residual LSTM, consisting of one LSTM layer between two linear layers.

\subsection{Data Collection}
\label{sec: Data Collection}
We collected training data for the offline experiments using a Franka Emika Panda robot. 
The robot joint angles and velocities are measured directly from the robot.
Further, we approximate the acceleration of the robot using the filtered velocity signal. 
For the filtering, we use a non-causal filter, by applying a causal Butterworth filter once forward and then backward to cancel the phase shift.
The filter has a cutoff frequency of 5 Hz.
For the reference signal, we use the desired torque $\boldsymbol{\tau}_{\text{ref}}$ computed by our stiff impedance controller; as opposed to the measured torque signal from the Panda robot, which internally performs friction compensation. 
\begin{equation}
    \boldsymbol{\tau}_{\text{ref}} = \boldsymbol{\tau}_{\text{fb}} + \boldsymbol{\tau}_{\text{ff}}
\end{equation}
For the data collection, we use an inverse dynamics PD controller with proportional gains of $[400, 400, 200, 200, 200, 200]$ and damping gains tuned for critical damping.
The controller uses a rigid-body inverse dynamics model, which has been optimized with SDP.
The training data set is obtained from optimal experiment design ~\cite{swevers1997optimal}, where trajectories designed to maximize the information gain and numerical conditioning of the regression problem obtained by stacking Eq. \eqref{eq:linear EOM} for many measurement samples. 
The data collection procedure is commonly used for system identification of rigid-body dynamics models and hence we name this the system identification data. 
These trajectories have several advantages for system identification, such as time-domain averaging and avoiding frequency leakage errors.
They try to cover large parts of the state-space by moving the robot with sine and cosine movements. 
The trajectories for the joint angles are described by
\begin{equation}
\label{eq: sys id trajectories}
    \begin{split}
    q_{i}(t) = \sum_{l=1}^{N} \frac{a_{l}^{i}}{\omega_{f}l} \sin(\omega_{f}lt) - \frac{b_{l}^{i}}{\omega_{f}l} \cos(\omega_{f}lt) + q_{0} \\
    \end{split}
\end{equation}
with $\omega_{f}$ the fundamental pulsation \cite{swevers1997optimal}.
Before optimizing the trajectories, the initial values for the parameters $a_{l}^{i}$, $b_{l}^{i}$, and $q_{0}$ are chosen randomly. Furthermore, $\omega_{f}$ is chosen sufficiently low, such that a large amplitudes can be achieved without violating velocity and acceleration bounds. On the other hand, $N$ is selected such that excitation of high-frequency modes is avoided.
Each of the 20 trajectories generated are executed in two manners. 
Once with a maximum velocity $0.1\boldsymbol{\dot{q}}_{\text{max}}$
and once faster with a maximum velocity of $0.4\boldsymbol{\dot{q}}_{\text{max}}$, where $\boldsymbol{\dot{q}}_{\text{max}}$ is as provided by the robot manufacturer.
All the signals are collected at 1000 Hz.
We generate 26 different trajectories by varying the seed of the random number generator used to initialize the optimizer and execute them in a slow and fast manner. In total, we collected around 5 hours of data, i.e., about 17.5 million data points.
The recorded data was then subsequently split into a separate training, validation, and test set containing 42, 6, and 4 recorded movements, respectively.
The test set consists of two trajectories executed in a slow and fast manner, which are not in the training or validation data. We stop training by monitoring the validation loss of the model. 
As a consequence, we decided to design the validation data to consist of two trajectories executed with different settings unseen in the training, and two trajectories which are included with a different execution speed in the training data.

\subsection{Offline Data Experiments}
\label{sec: offline}
To compare the offline model learning performance, all models are implemented in Pytorch \cite{paszke2019pytorch} and trained using the mean squared error (MSE) loss for the normalized torque predictions.
We use the same normalization for all torque values to guarantee a fair comparison of the results:
$\tau_{\text{scaled},k} = \frac{\tau_{k} - \tau_{\text{min},k}}{\tau_{\text{max},k} - \tau_{\text{min},k}} $,
where $\tau_{\text{min}, k}$ and $\tau_{\text{max}, k}$ respectively denote the minimum and maximum observed torque values across all recorded training data for joint index $k$.
The values for all $\tau_{\text{min}, k}$ and $\tau_{\text{max}, k}$ are summarized in Table \ref{tab: lstm hyper param}.
The offline experiments are split into 3 sections: first, the different methods for identifying inertia parameters are evaluated against each other. Second, we test different models for their ability to be learned in an end-to-end setting together with a residual neural network.
Finally, we evaluate our proposed hybrid model architecture against the current baseline models introduced in Section \ref{sec:baseline}.

\subsection{Inertial Parameter Identification}
\label{Inertial parameter identification}

\begin{table}
        \centering
        \begin{tabular}{lccccc}  
        \toprule
            &  Training Loss &   Validation Loss & Test Loss  \\
             \midrule
            SDP  &   \textbf{0.0070}$\pm$(0.0000) &   \textbf{0.0074}$\pm$(0.0000)  &   \textbf{0.0085}$\pm$(0.0000) \\
            DiffBary  &  0.0071$\pm$($10^{-4}$) & 0.0075$\pm$($10^{-4}$) &   0.0086$\pm$($10^{-4}$) \\
            DiffNEA  & 0.0071$\pm$($10^{-4}$)  &  0.0075$\pm$($10^{-4}$)  &   0.0087$\pm$($10^{-4}$) \\
            DeLaN  &  0.0081$\pm$($3 \!\!\times\!\! 10^{-4}$)  &  0.13066$\pm$(0.0834)  &  0.2718$\pm$(0.2259) \\
            \bottomrule
        \end{tabular}
        \caption{Comparison of the average normalized torque prediction NMSE [$-$] and its standard deviation of inertial parameters retrieved using SDP, DiffNEA, and DiffBary method and DeLaN on a small training data set and two independent validation and test sets. The inertial parameters identified with SDP have the lowest average error. DiffNEA and DiffBary have similar performances. DeLaN has a similar training loss to the other models. However, it is not able to generalize its performance to unseen data.}
        \label{tab: Comparison of RBD learning}
\end{table}
We compare the performance of the DiffBary method against rigid body dynamics baselines~\cite{sousa2019inertia} and other hybrid baselines~\cite{sutanto2020encoding,lutter2019deep} to identify a set of feasible inertial parameters. 
To do so, we train all models on a small data set consisting of 2 system identification trajectories, without the friction and residual components. These trajectories follow the same path but are executed in a slow and fast manner.
\par
The DiffBary, DiffNEA, and DeLaN models are trained in Pytorch using Adam optimizer with a batch size of 1000 and with a learning rate of 0.004 (0.001 for DeLaN). The DeLaN model is trained with two layers of 200 neurons. We stop the training of all models when no improvement in loss is observed after 10 consecutive epochs. 
We evaluate the models on the validation and test data set to analyze their generalization capabilities. The results of the torque prediction on the training and test datasets are shown in Table \ref{tab: Comparison of RBD learning}.
The hybrid formulations of DiffBary and DiffNEA achieve similar torque prediction errors as that of classical RBD models obtained by SDP, on both training and test datasets. Thus we conclude that the hybrid formulations of DiffBary and DiffNEA are reliable models for finding a set of feasible inertial parameters. 
We investigate this claim further using model control experiments in Section \ref{sec:Experiment Results motion tracking}.
As seen in Table \ref{tab: Comparison of RBD learning} the DeLaN model fails at generalizing to unseen data, though it achieves a similar prediction error on the training set.
Interestingly the methods to identify inertial parameters do not converge to the same solutions. 
We hypothesize, that this is explained by the different solvers that are used and to what accuracy they solve the regression problems. That is, the stochastic Adam optimizer in a mini-batch optimization scheme for the PyTorch model versus a (deterministic) interior-point method for the SDP.
An overview of the resulting base parameters of DiffBary and SDP method is summarized in Table \ref{tab: base param comparison} of the Appendix.

\subsection{Joint Structured Model Learning}
\label{sec:residual-learning}
We now compare the performance of different rigid-body dynamics formulations when learned jointly with the residual neural network weights as in Figure \ref{fig: structured model overview}, using gradient descent methods.
Therefore, DiffBary is tested against the DiffNEA and the DeLaN model in an end-to-end learning setting with the residual LSTM. 
We investigate if the joint learning correctly identifies the inertial parameters of the RBD model inside the hybrid model shown in Figure \ref{fig: structured model overview} without the friction part.
To do so, we record the torque prediction error by the RBD model component in addition to the torque prediction error from the hybrid model on the validation data set.
As a baseline, we train the individual rigid-body dynamic models separately on the same data set without the residual part.

We deploy the Adam optimizer with a learning rate of 0.004 (0.001 for DeLaN) with a batch size of 1000 and train all models on 10 random seeds to track the average performance.
All residual hybrid models with LSTMs are trained using small sequences of data with 100 time steps each.
The LSTM has an internal hidden state, which is updated during each time step in a sequence.
We set the hidden state to zero at the beginning of every sequence. 
Training the models on the smaller parts of the trajectories instead of the complete ones makes the training more stable.
The residual LSTM inside each hybrid model has the hyperparameters listed in Table \ref{tab: lstm hyper param} of the appendix.
The results of our experiment are visualized in Figure \ref{fig: joint learning comparison}.
The average validation loss of all model types is shown together with the 5 percentile as the lower bound and the 95 percentile as the upper bound. As seen in Figure \ref{fig: joint learning comparison} for both DiffNEA and our method, the joint learning of the RBD model with a residual LSTM does not affect the correct identification of a valid set of inertial parameters for the RBD model. 
For both DiffNEA and DiffBary, the RBD model converges to an optimal value quickly in a few epochs. 
After the first few epochs, the model learns to capture the non-rigid-body dynamics effects using the residual LSTM resulting in a further decrease in the hybrid loss. 
Thus our architecture correctly disentangles the RBD and residual effects in an interpretable manner.
We make a similar observation for DeLaN but with a much higher torque prediction error from both RBD and residual model components. We conclude that both DiffNEA and DiffBary are suited for joint end-to-end learning of a hybrid residual model. 
In the following experiments, we choose the DiffBary method, however, the DiffNEA method would perform equally well.

\subsection{Inverse Dynamics Learning Comparison}
\label{sec:Task-specific Inverse Dynamics Learning}

\begin{table*}
         
        \centering
        \begin{tabular}{lcccccc}  
        \toprule
        \multirow{2}{*}{Model Type}& \multirow{2}{*}{Optimization} &  \multicolumn{3}{c}{Evaluation} \\
        \cmidrule(lr){3-5}
        & &  Train Loss (NMSE) & Val Loss (NMSE) & Test Loss (NMSE) \\
        \midrule
        \multirow{2}{*}{RBD+Sign+LSTM} & E2E & 0.0021$\pm$($6  \!\!\times\!\! 10^{-4}$) & 0.0025$\pm$($10^{-4}$) &  \textbf{0.0033}$\pm$($4 \!\!\times\!\! 10^{-4}$)  \\
        & 2S   &  0.0013$\pm$($5 \!\!\times\!\! 10^{-4}$) & \textbf{0.0022}$\pm$($2  \!\!\times\!\! 10^{-4}$) & 0.0035$\pm$($3  \!\!\times\!\! 10^{-4}$) \\
        RBD+Sign+MLP & 2S   & \textbf{0.0004}$\pm$($10^{-4}$) & 0.0030$\pm$($3  \!\!\times\!\! 10^{-4}$) & 0.0065$\pm$($4  \!\!\times\!\! 10^{-4}$)  \\
        \midrule
        \multirow{2}{*}{RBD+Sign} & E2E & 0.0038$\pm$($10^{-4}$) & 0.0036$\pm$($10^{-4}$) & 0.0037$\pm$($10^{-4}$)  \\
         & 2S & 0.0035$\pm$(0.0000) & 0.0036$\pm$(0.0000) & 0.0035$\pm$(0.0000)  \\
        \multirow{2}{*}{RBD} & E2E & 0.0082$\pm$($2  \!\!\times\!\! 10^{-4}$) & 0.0076$\pm$($10^{-4}$) & 0.0082$\pm$($10^{-4}$)  \\
         & 2S & 0.0080$\pm$(0.0000) & 0.0075$\pm$(0.0000) & 0.0081$\pm$(0.0000)  \\
        LSTM &  E2E  & 0.0036$\pm$(0.0011) & 0.0056$\pm$($3  \!\!\times\!\! 10^{-4}$) & 0.0089$\pm$($9  \!\!\times\!\! 10^{-4}$) \\
        \bottomrule
        \end{tabular}
        \caption{
        Comparison of different inverse dynamics models average torque prediction accuracies on different data sets. The performance of the end-to-end trained (E2E) hybrid model is disentangled by showing the accuracy of the RBD model part and the RBD model with Sign friction(RBD+Sign) against the performance with the residual LSTM. The hybrid model trained in 2 steps (2S) has the best validation performance, while its version trained end-to-end (E2E) has the lowest average test loss. Extending the rigid-body dynamics model (RBD) with the sign friction function (RBD+Sign) boosts its average performance. The LSTM is not able to generalize to unseen data outside the vicinity of the training distribution.}
        \label{tab:compinvdyn}
\end{table*}
Now, we aim to learn a highly precise inverse dynamics model applicable for motion tracking with the real robot. 
We study different inverse dynamics models trained on the complete dataset and
 evaluate their offline torque prediction errors on a separate test data set detailed in Section \ref{sec: Data Collection}.
For all models, we apply early stopping, where the training is interrupted when the validation loss does not improve after 30 consecutive epochs. 
A learning rate scheduler is deployed to lower the learning rate by the factor of 0.1 after 30 epochs. 
All models are trained with a batch size of 1000 samples and the Adam optimizer with a learning rate of 0.004.
An overview of the hyperparameters of the training is summarized in Table \ref{tab: lstm hyper param} of the appendix.
For the evaluation, the models predict the whole trajectories in one large sequence.
This allows the LSTM models to use their internal memory throughout the whole trajectory.

\textbf{End-to-End vs 2 Step Learning:}
First, we analyze the capabilities of end-to-end learning against a separate system identification for the residual hybrid model, which will be referred to as the "2-steps" model. The training of the 2-steps model involves two steps. In the first step, inertial parameters are identified using SDP. In the second step, the residual weights of the LSTM are learned in a second optimization using gradient descent with the RBD model parameters fixed.
The performance of both model types is summarized in Table \ref{tab:compinvdyn}.
\begin{figure}
    \centering
    \includegraphics{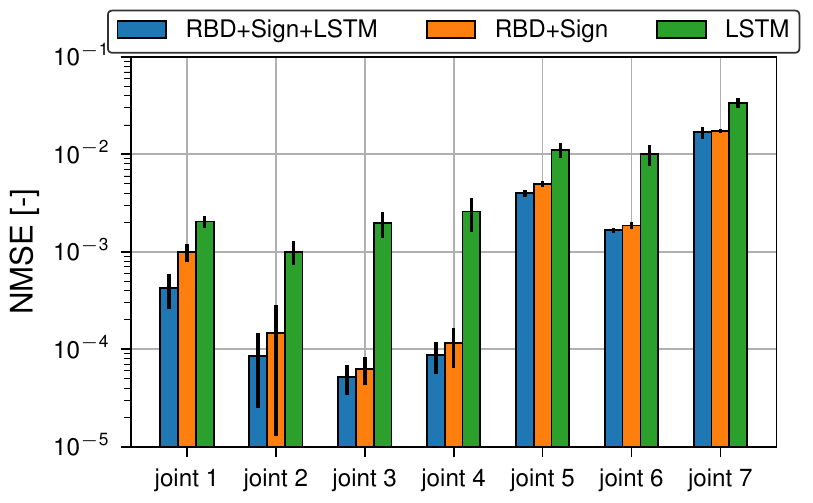}
    \caption{Comparison of our proposed residual hybrid model trained end-to-end (E2E), its rigid-body dynamics model with sign friction (RBD+Sign), and a black-box LSTM on the test data set consisting of four system identification trajectories. The average individual torque prediction NMSE for all joints is plotted for 10 different seeds. Both, the rigid-body dynamics model with friction, and the hybrid model outperform the black-box LSTM on all joints. The additional residual LSTM in combination with the RBD model boosts the average performance for all joints.}
    \label{fig: sysid data comparison}
\end{figure}
The results show that the 2 steps model has a lower training and validation accuracy, while the end-to-end model has a lower test error.
The average performance of the RBD model inside the end-to-end hybrid model (RBD E2E) confirms the results of Section \ref{sec:residual-learning}. 
Optimizing the barycentric weights alongside the neural network weights achieves similar accuracy as the RBD model optimized by SDP.

\par
\textbf{Residual LSTM vs MLP:}
Secondly, we compare two variants of the residual hybrid model, 1) with a residual LSTM and 2) with a residual MLP. We perform this ablation to evaluate the influence of the internal memory on the hybrid model's accuracy.
The torque prediction errors of Table \ref{tab:compinvdyn} show that the residual MLP achieves a lower torque prediction loss on the training data, while the LSTM outperforms the MLP on the validation and test loss.
Our ablation suggests that the LSTM fares better when it comes to generalization capabilities.
Thus introducing internal memory into the hybrid model improves the generalization accuracy at the cost of task-specific accuracy. 

\textbf{Hybrid vs Black-Box LSTM:} Finally, we compare our hybrid architecture with a black-box LSTM.
The individual torque prediction losses on the test data are visualized in Figure \ref{fig: sysid data comparison}.
The black-box LSTM achieves similar training loss while having a higher validation and test loss.
The residual hybrid model with the LSTM performs well on all datasets and is still more accurate than the baseline RBD model on unseen data, except for the last joint.
Comparing the bar plots of the hybrid model against the RBD shows, that the hybrid model achieves lower task-specific prediction error than its rigid-body baseline with friction while maintaining the good generalization performance of the RBD model.
The general improvement of the hybrid models over the RBD variants appears to be small in Table \ref{tab:compinvdyn} and Figure \ref{fig: sysid data comparison} of the offline evaluation. 
However, the real robot experimental results in Table \ref{tab: avrg motion tracking performance} show, that these small differences in the average torque prediction lead to large motion tracking errors.

\section{Motion Tracking Experiments}
\label{sec: tracking}
\begin{figure*}
    \centering
    \includegraphics{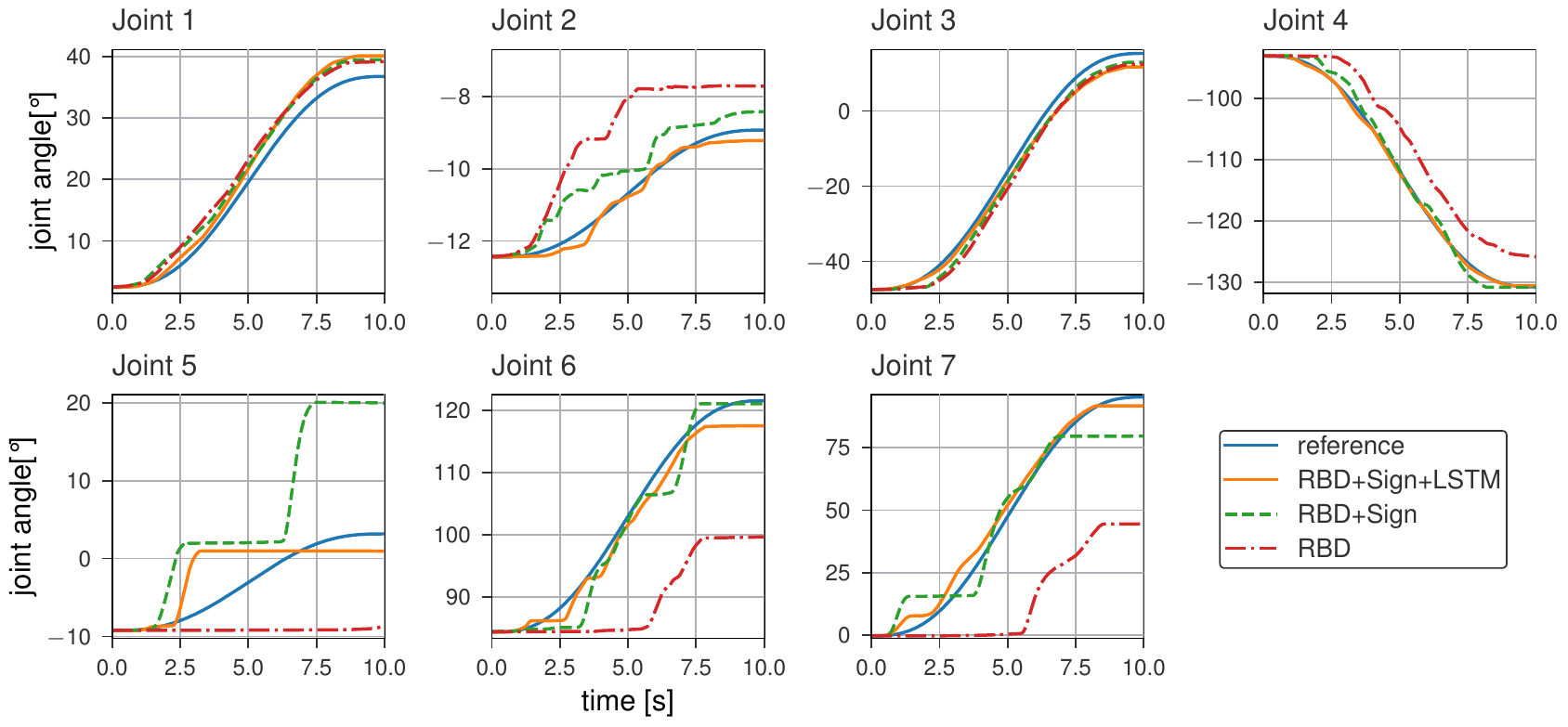}
    \caption{Motion tracking performance for a minimum-jerk trajectory of the proposed end-to-end learned hybrid model. The tracking performance is disentangled into 3 models by testing the rigid-body dynamics model (RBD) inside the hybrid model independently with sign friction and without. All models are combined with the medium controller settings on a 10 seconds movement. The results show, that the residual hybrid model is able to further increase the tracking accuracy for 5 out of 7 joints.}
    \label{fig: motion tracking comparison}
\end{figure*}
Motion tracking experiments evaluate the performance of inverse dynamics models and test their generalization abilities.
We deploy different inverse dynamics models inside a joint-space impedance-controlled Panda robot.
The inverse dynamics models are deployed as the feed-forward term inside the controller. 
The robot consecutively performs 5 minimum-jerk trajectories \cite{flash1985coordination}, which are generated from a set of joint state set-points.
The impedance controller imposes a desired dynamic behavior of a spring-mass-damper system for each generalized robot coordinate independently and uses the inverse dynamics to compute the corresponding reference torques.
A detailed description of the controller is given in Section \ref{sec: Joint Space impedance control} of the Appendix.

We evaluate the following models: an end-to-end (E2E) residual hybrid model, a hybrid model trained in 2 steps (2S), a residual hybrid model with an MLP, an RBD model optimized with SDP, and one RBD model identified inside the end-to-end learned hybrid model and a black-box LSTM.
All controller variants with an LSTM inside are initialized before starting the experiment. 
Therefore, the control loop is run 500 iterations to update the internal memory of the LSTM to the current system state.

\subsection{Experiment Details}
\begin{table*}[t]
	\centering%
	\begin{tabular}{lccccccccc}
	\toprule
	 & \multirow{2}{*}{Controller gains}&  \multicolumn{2}{c}{RBD+Sign+LSTM}  & RBD+Sign+MLP & \multicolumn{2}{c}{RBD+Sign} &  \multicolumn{2}{c}{RBD} & LSTM\\
	 \cmidrule(lr){3-4}\cmidrule(lr){5-5}\cmidrule(lr){6-7}\cmidrule(lr){8-9}\cmidrule(lr){10-10}
	 & & 2S & E2E & 2S & E2E & 2S &  E2E & 2S & E2E  \\
	\midrule  
    \multirow{3}{*}{\specialcell{Average joint angle \\RMSE[°]}} & Soft   & 16.207 & \textbf{12.799} & 22.442  & 22.892 & 18.372 & 19.013 & 18.867 & failed\\
    & Medium   & 2.599 & \textbf{2.244} & 3.719 & 2.948 & 6.764 &  11.613& 12.495& failed\\
    & Stiff   & 0.763 & \textbf{0.744} & 1.053 & 2.863 & 2.781  & 4.005 & 7.729 & failed\\
    \midrule  
   \multirow{3}{*}{\specialcell{Final joint angle \\offset[°]}} & Soft   & 24.45 & \textbf{18.401} & 34.427  & 30.053 & 24.379&  29.757 & 29.733& failed \\ 
    & Medium   & \textbf{2.304} & 3.125 & 3.955 &  3.447 & 9.075 &  16.388& 18.538& failed \\
    & Stiff   & \textbf{0.596} & 0.992 & 0.947 & 1.09 & 3.21 & 4.478 & 9.739& failed\\
    \bottomrule
	\end{tabular}
	\caption{Overview of the average joint angle tracking RMSE and the final joint angles error of different inverse dynamics models inside the impedance controller. All models are evaluated on 5 trajectories in a slow and fast manner. The results show, that the end-to-end hybrid model with the LSTM referred to as (E2E) outperforms all other models tested in the controller at 1000 Hz. Both RBD models identified end-to-end (E2E) with the DiffBary method and in two steps (2S) with the baseline SDP method have similar performance. The black-box LSTM model inside the controller immediately runs into the angle constraints of the joints. Hence we do not report results with black-box LSTM models.}
	\label{tab: avrg motion tracking performance}
\end{table*}
All trajectories are executed in slow and fast versions of 2 and 10 seconds to test the generalization performance of the different models to different velocities.
For the final evaluation we record the joint state desired pose $\boldsymbol{q}_{\text{des}}^{t}$ and current pose $\boldsymbol{q}^{t}$ during the execution at 1000 Hz.
The final position offset at the end of every movement is evaluated in addition to the significance of the final position error for practical applications.
We evaluate the performance of our models with three different controller gain settings and thereby illustrate the sensitivity of controller performance to these gains: Soft: $[1, 1, 1, 1, 1, 1, 1]$, Medium: $[50,50,50,50,50,50,50]$ and Stiff: $[400, 400, 200, 200, 200, 200]$ together with  damping constants tuned for critical damping. 
We remark that the  stiff set of gains is the preferred setting for most activities in the lab.
In the experiment with the lowest gains (Soft), which are two orders of magnitude lower than the stiffest (Stiff), the tracking performance is dominated by the trained model and the influence of the feedback controller is small. 
For the evaluation, we separately calculate the RMSE of the joint angles during movement, and the final position offset. 

\subsection{Experiment Results}
\label{sec:Experiment Results motion tracking}
The motion-tracking results of the different models are summarized in Table \ref{tab: avrg motion tracking performance}.
A visualization of the joint angle tracking performance on one sample sequence in Figure \ref{fig: motion tracking comparison} further illustrates the performance increase of the hybrid model over its baseline RBD model with sign friction using the example of a trajectory with a medium controller gains.

Contrary to the offline torque prediction results, where the hybrid model trained in 2 steps excels, the end-to-end hybrid model achieved the best performance overall.
The results support our hypothesis from the offline experimental results that recurrent models with an internal memory outperform stateless models like MLP.
This supports the initial argument, that internal memory is important for the accuracy of the model.
Embedding the RBD model into the recurrent model architecture allows the model to
compensate for lower feedback gains while still improving the overall tracking performance in comparison to all RBD models. 
Thus, rendering them ideal for compliant and precise control applications.

We also note that the RBD model from the end-to-end learned hybrid model has a similar tracking performance as the RBD baseline models identified with SDP.
Interestingly, the end-to-end learned RBD model with the sign friction function outperforms the SDP variant.
These results provide evidence, that the end-to-end learning of hybrid models correctly disentangles forces caused by inertia, friction, and other effects, which boost the individual performance of each model type. 
Thus, these models achieve the best generalization accuracy.

\section{Conclusion} 
\label{sec:conclusion}

We propose a new residual hybrid model for inverse dynamics of robot manipulators. 
It combines a rigid body dynamics model, a simple friction model, and a recurrent neural network.
This combination allows the model to rely on strong physical priors for improved generalization and enables it to capture partially observable dynamics effects, e.g.,  stick-slip friction, and joint and link flexibilities. 
Furthermore, based on a recent method~\cite{sousa2019inertia} for optimizing fully physical consist rigid body dynamics parameters and inspired by \cite{sutanto2020encoding}, we propose a model formulation suitable for unconstrained gradient-based optimization.
Both methods allow joint, end-to-end training of all parts of the hybrid model, which results in improved online tracking results, compared to a two-step approach of first learning the rigid body dynamics and then fitting the network to the remaining error.
Compared to rigid body dynamics, our hybrid approach shows improved accuracy, and compared to black-box methods it demonstrates better generalization capabilities. 
We deployed our model as the inverse dynamics map for a computed-torque impedance controller, controlling a real robot at a frequency of 1kHz. 
In this setup, the model allows us to significantly reduce the feedback gains of the controller while mitigating a loss of tracking accuracy, thus achieving more precise and compliant control.
In future work, we want to investigate how we can exploit the fast gradient-based adaptation of all parameters in scenarios with changing dynamics, such as robot grasping.

\section*{Acknowledgments}
The research that led to this paper was funded by Robert Bosch GmbH. 
This work was also supported by the Deutsche Forschungsgemeinschaft (DFG, German Research Foundation) – 448648559 and the EPSRC UK (project NCNR, National Centre for Nuclear Robotics, EP/R02572X/1).

\bibliographystyle{plainnat}
\bibliography{references}

\clearpage

\appendix

\subsection{Learning fully physically consistent barycentric parameters using unbounded non-linear regression}
\label{sec: barycentric parameters using unbounded non-linear regression}
The barycentric parameter equation is a linear mapping of the inertia parameters in the joint link frame to the torque.
\begin{equation}
    \label{simple bary equation}
    \boldsymbol{\tau}_{\text{ref}} = \boldsymbol{Y}(\boldsymbol{q}, \boldsymbol{\dot{q}}, \boldsymbol{\ddot{q}}) \boldsymbol{\theta}
\end{equation}
This equation can be optimized using simple gradient-descent methods.
However, there are no guarantees of the physical consistency of the inertia parameters. 
Link frame inertia parameters are fully physically consistent if the linear matrix inequality of Eq. \eqref{eq:LMI inertial paraeters} is kept for every joint link.
Writing out the equation yields
\begin{align}
	\label{LMI Sousa extended}
	\begin{split}
	\boldsymbol{S}_{k} & = \\ 
	& \left[\begin{array}{@{}c@{\shrink}c@{\shrink}c@{\shrink}c@{}c@{\hspace{-0.3em}}c@{}} 
	\frac{(L_{yy} + L_{zz} -L_{xx})}{2} & -L_{xy} & -L_{xz} & l_{x} \\ -L_{xy} & \frac{(L_{xx} + L_{zz} - L_{yy} )}{2} & -L_{yz}  & l_{y} \\ -L_{xz} & -L_{yz} &  \frac{(L_{xx} + L_{yy} - L_{zz})}{2} & l_{z}  \\ l_{x} & l_{y} & l_{z} & m \end{array} \right] \succ 0.
	\end{split}
\end{align}
The fully physical consistency inequality for any joint link is defined in dependency of all 10 inertia parameters.

We apply the idea of \cite{sutanto2020encoding} to use the terms of the Cholesky decomposition to guarantee a positive-definite matrix.
Therefore, we introduce a lower triangular matrix $\boldsymbol{A}_{k}$ of a generic link frame $k$, which is expressed in dependency of 10 unbounded parameters.
\begin{equation}
	\boldsymbol{A}_{k} = \begin{pmatrix} a_{k} & 0 & 0 & 0\\ e_{k} & b_{k} & 0 & 0\\ f_{k} & g_{k} & c_{k} & 0 \\ h_{k} & i_{k} & j_{k} & d_{k} \end{pmatrix}
\end{equation} 
Now the physical barycentric parameters are constructed from the positive definite matrix product $\boldsymbol{A}_{k}\boldsymbol{A}_{k}^{\top}$ of the lower triangular matrix $\boldsymbol{A}_{k}$ together with a small position bias $r$ for all diagonals.
\begin{align}
\label{eq: S from unbounded}
    \begin{split}
	\boldsymbol{S}_{k} & = \boldsymbol{A}_{k} \boldsymbol{A}_{k}^{\top} + r\boldsymbol{I}_{4}\\
     & =  \begin{bmatrix} \delta_1 & \delta_5 & \delta_6 & \delta_8 \\ \delta_5 & \delta_2 & \delta_7 & \delta_8\\\delta_6 & \delta_7 & \delta_3 & \delta_{10} \\ \delta_8 & \delta_9 & \delta_{10} & \delta_4 \end{bmatrix}.
\end{split}
\end{align}
where,
\begin{equation*}
\begin{aligned}
\delta_1 &= a_{k}^{2}  +r, \\
\delta_2 &= e_{k}^{2} + b_{k}^{2}+r, \\
\delta_3 &= f_{k}^{2}+g_{k}^{2}+c_{k}^{2}+r, \\
\delta_4 &= h_{k}^{2}+i_{k}^{2}+j_{k}^{2}+d_{k}^{2}+r, \\
\delta_5 &= a_{k}e_{k}, \\
\delta_6 &= a_{k}f_{k}\\
\delta_7 &= e_{k}f_{k} +g_{k}b_{k}, \\
\delta_8 &= a_{k}h_{k}, \\
\delta_9 &= c_{k}h_{k}+i_{k}b_{k}, \\
\delta_{10} &=f_{k}h_{k}+g_{k}i_{k}+c_{k}j_{k}
\end{aligned}
\end{equation*}
Now all the inertial parameters can be retrieved from the positive-definite matrix $\boldsymbol{S}_{k}$.
Obtaining the values for $L_{xx},L_{yy},  L_{zz}$ requires additional calculations,
\begin{align}
\label{eq: retrive bary params}
    \begin{split}
        L_{xx} & =  \frac{(L_{xx} + L_{zz} -L_{yy})}{2} +  \frac{(L_{xx} + L_{yy} -L_{zz})}{2}\\ & =  e_{k}^{2} + b_{k}^{2} + f_{k}^{2}+g_{k}^{2}+c_{k}^{2} +2r\\
        L_{yy} & =  \frac{(L_{yy} + L_{zz} -L_{xx})}{2} +  \frac{(L_{yy} + L_{xx} -L_{zz})}{2}\\ & =  a_{k}^{2} + f_{k}^{2}+g_{k}^{2}+c_{k}^{2}+2r\\
        L_{zz} & =   \frac{(L_{yy} + L_{zz} -L_{xx})}{2} +  \frac{(L_{xx} + L_{zz} -L_{yy})}{2} \\ & =  a_{k}^{2} + e_{k}^{2} + b_{k}^{2}+2r,
\end{split}
\end{align}
while other parameters can directly be reconstructed
\begin{align}
    \begin{split}
        L_{xy} & =  -  a_{k}e_{k} \\
        L_{xz} & = - a_{k}f_{k} \\
        L_{yz} & =  - e_{k}f_{k}- g_{k}b_{k} \\
        l_{x} & =  a_{k}h_{k} \\
        l_{y} & = c_{k}h_{k} + i_{k}b_{k}\\
        l_{z} & = f_{k}h_{k} + g_{k}i_{k} + c_{k}j_{k} \\
        m & =  h_{k}^{2}+i_{k}^{2}+j_{k}^{2}+d_{k}^{2} +r.
\end{split}
\end{align}
For a 7 degree of freedom robot its barycentric parameters are identified from scratch using $7$ sets with $10$ virtual unbounded parameters $a_{k},...,h_{k}$, one for every joint $k$.
For our implementation we initialize all unbounded virtual parameters with a mean of $0$ and a standard deviation of $0.001$.

\subsection{Joint space impedance control}
\label{sec: Joint Space impedance control}

For the motion tracking experiments a classic joint space impedance controller is used~\cite{SiciBook}.
The impedance controller mimics the dynamics of mass-spring-damper system with adjustable parameters according to
\begin{equation}
	\label{joint impedance_equation}
	\boldsymbol{M}_{\text{des}}(\boldsymbol{\ddot{q}}^{t} - \boldsymbol{\ddot{q}}_{\text{des}}^{t}) + \boldsymbol{B}_{\text{des}}(\boldsymbol{\dot{q}}^{t} - \boldsymbol{\dot{q}}_{\text{des}}^{t}) + \boldsymbol{K}_{\text{des}}(\boldsymbol{q}^{t}-\boldsymbol{q}_{\text{des}}^{t}) =\boldsymbol{\tau}_{\text{ext}}^{t},
\end{equation}
where $\boldsymbol{M}_{\text{des}}$ is the desired inertia matrix, $\boldsymbol{B}_{\text{des}}$ the virtual dampening matrix, $\boldsymbol{K}_{\text{des}}$ denotes the spring constant matrix and $\boldsymbol{\tau}_{\text{ext}}$ are the joint torques caused by external forces.
We use feedback-linearization in joint acceleration space and insert Eq. \eqref{joint impedance_equation} into the inverted manipulator dynamics from Eq. \eqref{eom}. This yields the following torque control feedback law

\begin{eqnarray*}
\label{joint impedance_equation control equation}
	\boldsymbol{\tau}_{\text{ref}}^{t} &=& \boldsymbol{V}(\boldsymbol{q}^{t}, \boldsymbol{\dot{q}}^{t})  + \boldsymbol{G}(\boldsymbol{q}^{t})-\boldsymbol{\tau^t}_{\text{ext}}\\ &+&\boldsymbol{M}(\boldsymbol{q}^t)\boldsymbol{M}^{-1}_{\text{des}}\left(\boldsymbol{\tau}^t_{\text{ext}}-\boldsymbol{B}_{\text{des}}(\boldsymbol{\dot{q}}^{t} - \boldsymbol{\dot{q}}_{\text{des}}^{t}) - \boldsymbol{K}_{\text{des}}(\boldsymbol{q}^{t}-\boldsymbol{q}_{\text{des}}^{t})\right)\\
	&+&\boldsymbol{M}(\boldsymbol{q})\ddot{\boldsymbol{q}}^t_{\text{des}},
\end{eqnarray*}
with $\boldsymbol{\tau}^t_{\text{ref}}$ being the actuation torque. In case of free space motion, where $\boldsymbol{\tau}^t_{\text{ext}} = \boldsymbol{0}$, the above control law simplifies to classical inverse dynamics PD control.

\begin{table*}
        \label{tab: lstm hyper param}
        \centering
        \begin{tabular}{l|cccc}
        \toprule
         & \specialcell{Residual \\LSTM} & \specialcell{Residual \\MLP} & \specialcell{Black-box\\ LSTM} & DeLaN \\
        \midrule
    Input dimension       &  21  &  21&  21 & 21 \\
    Output dimension       &  7 &  7 &  7&  7\\
    Number LSTM layers      &  1 &  -&  1 & - \\
    LSTM hidden dimension      & 50 & -& 50 & - \\
    Number linear layers          & 1 & 2 & 1  & 2\\
    Linear dimension      & 100 & 200 & 100 & 200 \\
    Linear activation     & PReLU  & PReLU & PReLU & SoftPlus\\
    Number of weights &33358 & 46008 & 33358 & 47629 \\
    Time steps     & 100  & 1  & 100 & 100 \\
    Batch size     & 1000 & 1000 & 1000 & 1000 \\
    Learning rate     &  0.004   &  0.004 &  0.004 & 0.001\\
    \midrule
    \multicolumn{5}{c}{Inverse Dynamics Learning Comparison}\\
    \midrule
    Training samples   & \multicolumn{4}{c}{[135210, 100, 21]} \\
    Validation samples &\multicolumn{4}{c}{[25930, 100, 21]} \\
    Min torque values [Nm] &\multicolumn{4}{c}{[ -10.00, -63.80, -28.24, -24.98, -3.95, -4.21, -1.36]} \\
    Max torque values [Nm] &\multicolumn{4}{c}{[ 9.78, 61.37, 27.64, 27.9, 3.94, 3.84, 1.84]} \\
        \bottomrule
        \end{tabular}
        \caption{Overview of the optimized hyperparameters of the different residual and black-box LSTMs,  the DeLan model and the torque values used for normalization.}
\end{table*}


\begin{table*}
\label{tab: base param comparison}
\centering

\begin{tabular}{lccc}
\toprule
 Base Parameter & DiffBary & SDP  \\
 \midrule
 \rowcolor{lightgray}
 $L_{1yy} + L_{2zz}$ & 0.14240 &  0.05440  \\
 $L_{2xx} - L_{2zz} + L_{3zz} + \frac{79}{125}l_{3y} + \frac{372199}{4000000}(m_3 + m_4+m_5+m_6+m_7)$ & 1.21090 &  1.71400  \\
  \rowcolor{lightgray}
 $L_{2xy}$ & -0.00560 & -0.00040 \\
 $ L_{2xz}$ & 0.02060 &  0.03450 \\
  \rowcolor{lightgray}
 $L_{2yy} + L_{3zz} + \frac{79}{125}l_{3y} + \frac{372199}{4000000}(m_3 + m_4+m_5+m_6+m_7)$ & 1.22020 &  1.72490 \\
 $L_{2yz}$ & -0.06220 &  0.00600  \\
 \rowcolor{lightgray}
$l_{2x}$ & -0.05690 & -0.05410 \\
$l_{2z} + l_{3y} + \frac{79}{250}(m_3+ m_4+m_5+m_6+m_7)$ & 3.44070 &  3.42250  \\
\rowcolor{lightgray}
$ L_{3xx} - L_{3zz} + L_{4zz} + \frac{1089}{160000}m_3$ & 0.00860 & -0.00890  \\
$ L_{3xy} - \frac{33}{400}l_{3y}$ & -0.01050 & -0.14800 \\
\rowcolor{lightgray}
$L_{3xz}$ & -0.00010 &  0.00230 \\
$L_{3yy} + L_{4zz} - \frac{1089}{160000}(-m_3 - 2m_4 -2m_5-2m_6-2m_7 )$ & -0.08700 & -0.02990  \\
\rowcolor{lightgray}
$L_{3yz}$ & -0.01160 &  0.00570  \\
$l_{3x} + \frac{33}{400}(m_3+ m_4+m_5+m_6+m_7) $ &  0.84560 &  0.81150  \\
\rowcolor{lightgray}
$l_{3z} - l_{4y}$ & -0.00840 & -0.01160  \\
$ L_{4xx} - L_{4zz} + L_{5zz} + \frac{6}{125}l_{5y} + \frac{1089}{160000}m_4 + \frac{617049}{4000000}(m_5+m_6+m_7)$ &  0.76810 &  0.82510 \\
\rowcolor{lightgray}
$L_{4xy} + \frac{33}{400}l_{4y}$ & 0.00090 & -0.00370  \\
$ L_{4xz}$ & -0.00010 &  0.01910  \\
\rowcolor{lightgray}
$L_{4yy} + L_{5zz} + \frac{96}{125}l_{5y} - \frac{1089}{160000}m_4 + \frac{562599}{4000000}(m_5 + m_6+m_7)$ &  0.68210 &  0.77540  \\
$L_{4yz}$ & -0.00290 &  0.00110 \\
\rowcolor{lightgray}
$ l_{4x} - \frac{33}{400}(m_4+ m_5 + m_6 + m_7) $ & -0.52140 & -0.51670 \\
$l_{4z} + l_{5y} + \frac{48}{125}(m_5  + m_6 +m_7)$ &  2.05570 &  2.05900  \\
\rowcolor{lightgray}
$L_{5xx} - L_{5zz} + L_{6zz} - \frac{121}{15625}(m_6 + m_7)$ &  0.03670 & -0.00180  \\
$L_{5xy}$ & -0.00090 &  0.01240 \\
\rowcolor{lightgray}
$L_{5xz}$ &  0.00040 &  0.00080  \\
$L_{5yy} + L_{6zz} - \frac{121}{15625}(m_6 + m_7)$ &  0.03690 &  0.02670 \\
\rowcolor{lightgray}
$L_{5yz}$ &  0.00530 &  0.00040  \\
$l_{5x}$ &  0.00710 &  0.00610  \\
\rowcolor{lightgray}
$l_{5z} + l_{6y}$ & -0.08870 & -0.08830 \\
$ L_{6xx} - L_{6zz} + L_{7yy} +  \frac{107}{500}l_{7z} + \frac{121}{15625}m_6 + \frac{19193}{1000000} m_7$ &  0.00990 &  0.01460 \\
\rowcolor{lightgray}
$L_{6xy} - \frac{11}{125}l_{6y}$ & -0.01520 & -0.00770  \\
$L_{6xz}$ & -0.01760 & -0.00030  \\
\rowcolor{lightgray}
$L_{6yy} + L_{7yy} + \frac{107}{500}l_{7z} - \frac{121}{15625}m_6 + \frac{741}{200000}m_7$ &  0.03790 &  0.01550 \\
$L_6yz$ & -0.00680 &  0.00050 \\
\rowcolor{lightgray}
$l_{6x} + \frac{11}{125}( m_6+ m_7)$ &  0.22110 &  0.22340  \\
$l_{6z} + l_{7z} + \frac{107}{1000}*m_7$ &  0.16480 &  0.16500  \\
\rowcolor{lightgray}
$L_{7xx} - L_{7yy}$  &  0.00020 & -0.00090  \\
$ L_{7xy}$ & -0.00040 & -0.00060  \\
\rowcolor{lightgray}
$L_{7xz}$ & -0.00060 &  0.00180  \\
$ L_{7yz}$ & -0.00070 &  0.00100 \\
\rowcolor{lightgray}
$L_{7zz}$ &  0.00090 &  0.00430  \\
$ l_{7x}$ &  0.00720 &  0.00860  \\
\rowcolor{lightgray}
$l_{7y}$ &  0.00950 &  0.00990  \\
\hline
\end{tabular}
\caption{Comparison of the identified Dynamic base parameters identified using SDP \cite{sousa2019inertia} and our new proposed DiffBary method}
\end{table*}

\begin{table*}
	\label{tab: final error of hybrid models with gains}
	\centering
	\begin{tabular}{lc|cccccccc}
    \toprule
    \multicolumn{9}{c}{Final Joint Angle Error with Soft Gains}\\
    \midrule
    & & joint 1  &  joint 2 &  joint 3 & joint 4  &  joint 5 &  joint 6 &  joint 7 \\
    \midrule
    RBD+LSTM+Sign& 2S &12.654& 20.557& 12.039&  8.238& 29.701& 19.135& 68.824 \\
    RBD+LSTM+Sign & E2E   &\textbf{10.245}& 10.879&  \textbf{6.773}& 14.452& 39.852& 11.664& 34.94 \\
    RBD+MLP+Sign & 2S & 31.706& {31.421}& {54.849}& 19.944& 26.753& 16.283& 60.034 \\
    RBD+Sign & E2E & 29.429& 16.983& 30.662& 13.045& {63.4}  & 12.895& 43.959 \\
    RBD+Sign & 2S & {31.231}& 11.852& 32.43 &  \textbf{9.26} & 49.03 &  \textbf{7.574}& \textbf{29.273} \\
    RBD & E2E &  25.796&  9.814& 21.706& {23.595}& 12.334& {24.934}& {90.121} \\
    RBD & 2S &20.5  &  \textbf{8.435}& 28.757& 23.062& \textbf{12.33} & 24.929& {90.121} \\
    \midrule
    \multicolumn{9}{c}{Final Joint Angle Error with Medium Gains}\\
    \midrule  
    RBD+LSTM+Sign  & 2S  &\textbf{1.145}& 1.115& \textbf{1.144} &  1.13& 5.575& 2.41 & \textbf{3.608} \\
    RBD+LSTM+Sign & E2E &1.855& \textbf{0.399}&  1.82& \textbf{0.585} & 8.411& \textbf{1.184} & 7.621 \\
    RBD+MLP+Sign & 2S  & 1.338& 1.057& 1.68 & 1.062& 11.609& 4.032& 6.906 \\
    RBD+Sign & E2E & 3.561 & 0.993 & 3.81 & 0.903 & \textbf{3.049}  & 1.734 & 10.077 \\
    RBD+Sign & 2S & 4.409 & 0.69 & 4.904 & 1.101 & 4.084 & 1.84 &  46.497 \\
    RBD  & E2E &  3.701&  0.726& 3.486& 4.02& 12.069& 22.522& 68.194 \\
    RBD & 2S   & 3.576  &  0.697 & 3.134 & 3.177 & 12.303 & 16.831 & 90.045 \\
    \midrule
    \multicolumn{9}{c}{Final Joint Angle Error with Stiff Gains}\\
    \midrule  
    RBD+LSTM+Sign & 2S  &\textbf{0.119}& 0.323& \textbf{0.321}& 0.365& 1.152& 0.66 & \textbf{1.235} \\
    RBD+LSTM+Sign & E2E    &0.245& \textbf{0.15} & 0.504& \textbf{0.212}& 3.244& \textbf{0.24 }& 2.351 \\
    RBD+LSTM & 2S    &0.379& {0.514}& 0.34 & {1.677}& {6.586}& 3.857& 9.618 \\
    RBD+MLP+Sign & 2S  & 0.164& 0.193& 0.309& 0.431& 2.62 & 0.865& 2.046 \\
    RBD+Sign & E2E &0.434& 0.241& 0.923& 0.244& \textbf{0.787}& 0.741& 4.262 \\
    RBD+Sign & 2S   &{0.54 }&  0.177&  {1.228}&  0.343&  2.379&  0.92 & 16.884 \\
    RBD  &   E2E&  0.322&  0.18 &  0.6  &  1.067&  3.76 &  {6.084}& 19.331 \\
    RBD  & 2S & 0.399&  0.105&  0.867&  0.871&  4.881&  4.437& {56.613} \\
    \bottomrule
	\end{tabular}
	\caption{Comparison of final joint angles offset averaged for all motion tracking  trajectories for all different hybrid models with three different controller configurations}
\end{table*}

\begin{table*}
	\label{tab: average mean squared error of hybrid models with soft gains}
	\centering
	\begin{tabular}{lc|cccccccc}
	\toprule
    \multicolumn{9}{c}{Average Root Mean Squared Error with Soft Gains}\\
    \midrule
	&  & joint 1  &  joint 2 &  joint 3 & joint 4  &  joint 5 &  joint 6 &  joint 7 \\
	\midrule  
	RBD+LSTM+Sign & 2S    & 9.175& 13.139&  8.47 &  \textbf{5.904}& 17.176& 13.621& 45.967 \\
    RBD+LSTM+Sign  & E2E   & 5.358&  \textbf{6.347}&  \textbf{4.024}& 10.419& 26.368& 10.98 & \textbf{26.096} \\
	RBD+MLP+Sign  & 2S   & 19.259& {18.881}&{ 33.772}& 12.486& 27.59 & \textbf{ 9.733}& 35.372 \\
    RBD+Sign & E2E  & 19.492& 10.012& 20.51 &  9.06 & {43.468} & 10.253& 47.449 \\
    RBD+Sign  & 2S     & {20.816}&  6.578& 21.384&  6.751& 33.836&  8.944& 30.294]  \\
    RBD & E2E    & 15.483&  8.643& 14.473& {14.75} &  7.72 & {15.607}& {56.413} \\
    RBD  & 2S    & \textbf{3.012}&  6.75 & 17.876& 14.698&  \textbf{7.718}& 15.605& {56.413} \\
    \midrule
    \multicolumn{9}{c}{Average Root Mean Squared Error with Medium Gains}\\
    \midrule
    RBD+LSTM+Sign & 2S    &\textbf{1.367}& 0.779& \textbf{1.262}& 1.105& 4.811& 2.064& 6.802\\
    RBD+LSTM+Sign & E2E    &1.45 & \textbf{0.379}& 1.521& \textbf{0.621}& 5.666& 1.668& \textbf{4.401} \\
    RBD+MLP+Sign & 2S  & 2.088&  0.751&  1.905&  1.141& 10.354&  2.822&  6.971 \\
    RBD+Sign & E2E  & 3.191& 0.928& 3.46 & 0.925& \textbf{2.41} & \textbf{1.62} & 8.102 \\
    RBD+Sign   & 2S   & {3.991}&  0.693& { 4.42} &  1.068&  4.936&  1.722& 30.518 \\
    RBD   & E2E   &  3.656&  0.852&  3.487&  3.031&  7.574& 14.766& 47.923 \\
    RBD & 2S    & 3.596&  0.653&  3.339&  2.935&  7.702& 12.865& {56.374} \\
    \midrule
    \multicolumn{9}{c}{Average Root Mean Squared Error with Stiff Gains}\\
	\midrule
    RBD+LSTM+Sign & 2S   &\textbf{0.194}& 0.241& \textbf{0.372}& 0.398& 1.512& 0.701& 1.923\\
    RBD+LSTM+Sign & E2E    &0.293& \textbf{0.13} & 0.517& \textbf{0.187}& 2.277& \textbf{0.414}& \textbf{1.392} \\
    RBD+LSTM  & 2S & 0.539& {0.403}& 0.728& {2.008}& {9.278}& 6.34 & 7.109\\
    RBD+MLP+Sign & 2S & 0.319& 0.219& 0.544& 0.345& 3.054& 0.711& 2.181 \\
    RBD+Sign & E2E & 0.466& 0.272& 1.017& 0.276& \textbf{0.77} & 0.643& 3.176 \\
    RBD+Sign & 2S   & {0.593}&  0.225&  {1.354}&  0.316&  2.698&  0.757& 14.096 \\
    RBD  & E2E  &   0.529&  0.284&  1.006&  1.109&  2.915&  {5.575}& 16.62 \\
    RBD  & 2S  &0.545&  0.193&  1.058&  1.027&  3.997&  4.417& {42.864}\\
    \bottomrule
	\end{tabular}
	\caption{Comparison of root mean squared errors (RMSE) for all different hybrid models with all three controller configurations averaged for all motion tracking experiments}
\end{table*}

\end{document}